\documentclass[runningheads]{llncs}
\usepackage[T1]{fontenc}
\usepackage{booktabs}
\usepackage{multirow}
\usepackage[table]{xcolor}  

\usepackage{amssymb}
\usepackage{amsmath}
\usepackage{pifont}
\usepackage{graphicx}

\begin{document}
%
\title{Probing Fine-Grained Action Understanding and Cross-View Generalization of  Foundation Models}

%
\titlerunning{Probing Fine-Grained Action Understanding of Foundation Models}
%
\author{Thinesh Thiyakesan Ponbagavathi\inst{1} \and
Kunyu Peng\inst{2} \and
Alina Roitberg\inst{1}}
\authorrunning{Thiyakesan Ponbagavathi et al.}
%
\institute{Institute for Artificial Intelligence, University of Stuttgart, Germany \and
Institute for Anthropomatics and Robotics (IAR), KIT, Germany\\
\email{thinesh.thiyakesan-ponbagavathi@ki.uni-stuttgart.de, \\ kunyu.peng@kit.edu,alina.roitberg@ki.uni-stuttgart.de}}
\maketitle              
\begin{abstract}

Foundation models (FMs) are large neural networks trained on broad datasets, excelling in downstream tasks with minimal fine-tuning. 
Human activity recognition in video has advanced with FMs, driven by competition among different architectures. 
However, high accuracies on standard benchmarks can draw an artificially rosy picture, as  they often overlook real-world factors like changing camera perspectives. 
Popular benchmarks, mostly from YouTube or movies, offer diverse views but only coarse actions, which are insufficient for use-cases needing fine-grained, domain-specific actions. Domain-specific datasets (e.g., for industrial assembly) typically use data from limited static perspectives.

This paper empirically evaluates how perspective changes affect different FMs in fine-grained human activity recognition. We compare multiple backbone architectures and design choices, including image- and video-based models, and various strategies for temporal information fusion, including commonly used score averaging and more novel attention-based temporal aggregation mechanisms. 
This is the first systematic study of different foundation models and specific design choices for human activity recognition from unknown views, conducted with the goal to provide guidance for backbone- and temporal fusion scheme selection.
Code and models will be made publicly available to the community.

\end{abstract}
\section{Introduction}
Fine-grained Human Activity Recognition (HAR) is essential for a wide range of applications, such as human-robot interaction during industrial assembly~\cite{IKEA_ASM,sener2022assembly101largescalemultiviewvideo}, driver activity monitoring~\cite{drive_and_act}, and sports \cite{shao2020finegymhierarchicalvideodataset}.
Unlike general HAR, which identifies broad categories like \textsl{eating} or \textsl{playing football}, fine-grained HAR distinguishes between subtle actions, such as \textsl{opening} vs. \textsl{closing a bottle} or \textsl{attaching} vs. \textsl{aligning} the manipulated object. 
General HAR is usually evaluated on Youtube- or movie-based datasets~\cite{carreira2017quo} naturally featuring diverse viewpoints.
Domain-specific datasets for fine-grained HAR, on the other hand, rely on manual data collection with fixed camera setups~\cite{drive_and_act,IKEA_ASM}. In real-world scenarios, cameras are placed at varying angles, and models trained on static viewpoints tend to be less effective under these conditions \cite{junejo2008cross,wang2014cross}. 
This limitation underscores the need for models that can generalize well across diverse viewpoints.

\begin{figure}[t]
\vspace{-0.5em} 
    \centering
    \includegraphics[width=1\textwidth]{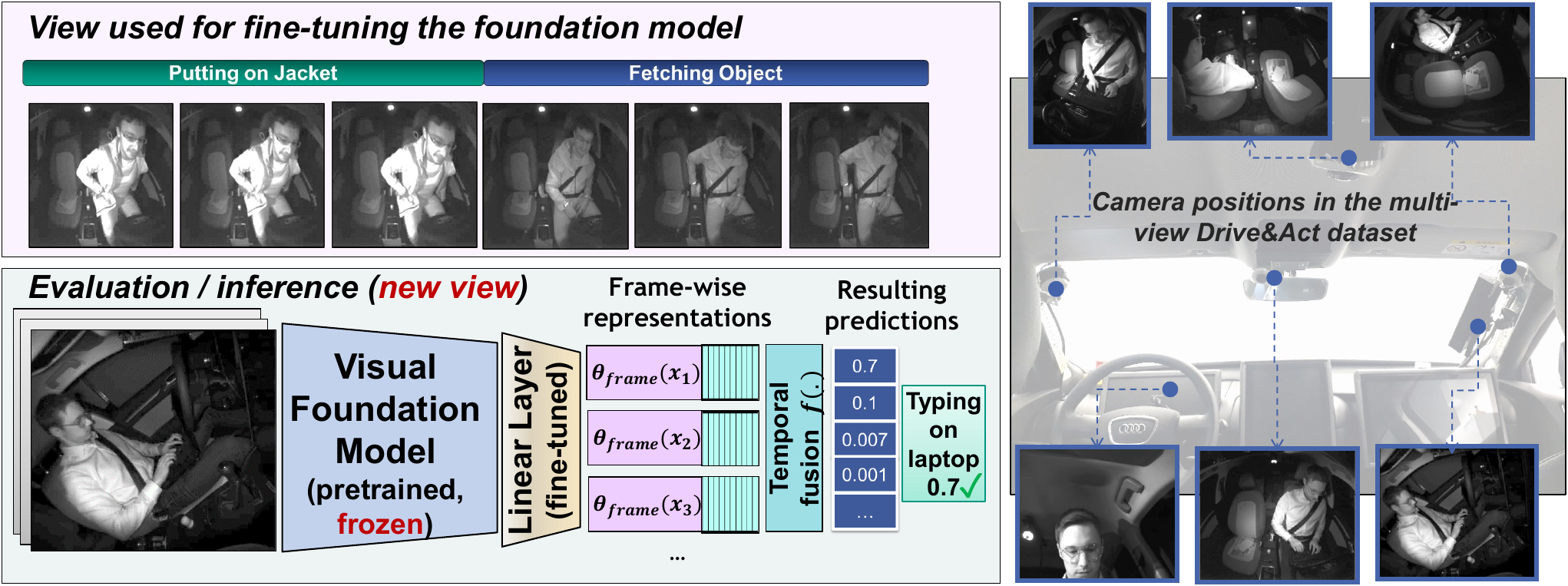}
    \caption{A high-level overview of a cross-view human activity recognition framework featuring pretrained frozen Foundation Models (FM) with linear probing and a temporal fusion mechanism. We study fine-grained activity understanding and cross-view generalization of different image- and video-based FMs and implement different techniques for linking temporal frame-level information.}
    \label{fig:teaser}
    \vspace{-1.5em} 
\end{figure}
Foundation models \cite{bommasani2021opportunities,CLIP,dinov2}, have shown promise in addressing similar generalization challenges. Pre-trained on extensive and diverse datasets in a self-supervised manner, these models learn generalized representations that can be fine-tuned for various downstream tasks, setting new standards for state-of-the-art (SOTA) performance. 
However, their understanding of \textit{fine-grained} activities and generalization to new views in particular remains under-explored. 

Given the great success of foundation models in diverse tasks, including general HAR~\cite{x-clip,tong2022videomae}, we ask whether they also represent fine-grained activities and deal with changes in camera perspective?
An important question arises from several architecture choices, such as video vs. image-based architectures and temporal fusion strategies for the latter group of methods, which has been overlooked in the past.
We conducted a series of experiments featuring different variants of four popular foundation models and found out that the granularity of actions impacts the performance of the foundation models. Among the foundation models utilized in this work, DinoV2 performs the best in all our benchmarks. We also implement 13 different mechanisms for linking temporal information, including standard averaging approaches as well as both published and new attention-based fusion mechanisms. We discover that self-attention based fusion mechanisms help image foundation models to outperform video foundation models in fine-grained HAR in both same-view and novel-view settings. 



\noindent The contributions of this work are summarized as follows:
\begin{itemize}
\item We for the first time consider linear probing of pretrained foundation models for fine-grained human activity recognition usually addressed with task-specific Transformers~\cite{liu2021video_swin,peng2022transdarc} and CNNs~\cite{aganian2023object,qiu2017learning,carreira2017quo}.
Our study is conducted on two datasets and features four foundation models with different design choices for fine-grained HAR with the goal of providing guidance for backbone- and temporal fusion scheme selection.
    \item We implement and systematically evaluate various temporal fusion techniques for adapting image foundation models to the video classification task. We show that image-based foundation models equipped with self-attention-based temporal fusion surpasses video foundation models.
    \item We also study cross-view generalization of foundation models. While preforming much better than the random baseline, perspective changes strongly affect the recognition quality, marking a clear need for research. Also in this setting, choosing the right temporal fusion mechanism is critical.
    
\end{itemize}

\section{Related Work}

\subsection{Foundation Deep Learning Architectures}
Foundation models are extensive machine learning models trained on large, diverse datasets, making them adaptable for numerous downstream tasks~\cite{luo2022clip4clip,li2022blip,li2023blip}. 
CLIP is proposed by Radford~\textit{et al.}~\cite{CLIP}, which verifies that state-of-the-art computer vision systems can be enhanced by pre-training on raw text-image pairs.
XCLIP~\cite{x-clip} is a video-based extenion of CLIP, which is a novel multi-grained contrastive model optimized for video-text retrieval.
BLIP is proposed by Li~\textit{et al.}~\cite{li2022blip} to bootstrap the efficacy of vision language model by using filtered web textual annotation. 
DinoV2 is a vision-only foundation model~\cite{dinov2}, which demonstrates that self-supervised pretraining on a large, curated, and diverse dataset can produce versatile visual features. 
Tong~\textit{et al.}~\cite{tong2022videomae} proposed VideoMAE by using masked autoencoders to achieve data-efficient learner for model pretraining. In this paper, we will explore the the ability of four popular foundation models (VideoMAE, CLIP, XCLIP, and DinoV2) to capture fine-grained human activities and deal with changes in camera perspective. 

\subsection{Fine-grained and Cross-View Human Activity Recognition}

Driven by advancements in general computer vision,  image-based~\cite{CLIP,dinov2} and video-based foundation~\cite{x-clip,tong2022videomae,v-jepa} models have been successfully employed for standard HAR, but \textit{fine-grained} activity recognition has seen less foundation model research. 
Examples of fine-grained HAR datasets include Drive~\&Act~\cite{drive_and_act}, IKEA-ASM~\cite{IKEA_ASM}, and FineGym~\cite{shao2020finegymhierarchicalvideodataset}.The slow progress of foundation models in this area motivates our work to explore their potential in fine-grained HAR.


Regarding HAR in new views, reformulated evaluation settings for several multi-view datasets, such as NTU \cite{shahroudy2016ntu,liu2019ntu}, have advanced research in this field, but the amount of research remains highly limited.
Skeletal pose-based methods are preferred for multi-view action recognition ~\cite{duan2022revisiting,chen2021channelwisetopologyrefinementgraph,BIAN2023103655} due to their accurate motion representation and reliable ground truth data. 
However, for \textit{fine-grained} activity recognition, RGB-based methods are preferred because skeletal poses lack the contextual information and subtle visual cues necessary for understanding \textit{fine-grained} actions. Recent RGB-based works \cite{siddiqui2023dvanetdisentanglingviewaction,multi-view_contrastive,viewclr} use a contrastive learning objective to learn view invariant representations  and achieve SOTA in cross-view HAR, but cross-view generalization for fine-grained human activity recognition is mostly uncharted. For example,  Drive\&Act~\cite{drive_and_act} and TransDARC \cite{peng2022transdarc} provide results of cross-view evaluation of standard action recognition CNNs and Transformers respectively, reporting a worrying drop in the recognition quality. In this work, we build on the evaluation protocols of the Drive\&Act~\cite{drive_and_act} and IKEA ASM~\cite{IKEA_ASM} benchmarks. The fine-grained nature of human activities recorded from multiple diverse views makes them an ideal testbed for validating fine-grained activity understanding and assessing the generalization of foundation models to changes in perspective.

\section{Revisiting Foundation Models for Cross-view Human Activity Recognition}
\subsection{Problem Statement}

We are primarily interested in studying foundation models for fine-grained human activity recognition with particular emphasis on robustness against changes of camera perspective.
Let $\mathcal{D}_{train}$ be the training dataset with videos captured from a fixed set of viewpoints $\mathcal{V}_{train}$.
At test-time, our model is exposed to new unseen viewpoints $\mathcal{V}_{novel}$, with $\mathcal{V}_{train} \cap  \mathcal{V}_{novel} = \emptyset $.
Given a sequence of frames $\mathbf{x} = \{x_1, \ldots, x_{t}\}$ captured from $\mathcal{V}_{test}$, the objective is to assign the correct action label $y$. 
This is achieved using a pretrained video classifier $\theta_{\text{clip}}$ or a combination of an image-based classifier $\theta_{\text{frame}}$ and a frame fusion function $f$. Both $\theta_{\text{clip}}$ and $\theta_{\text{frame}}$ are pretrained transformer-based foundation models, fine-tuned on $\mathcal{D}_{\text{train}}$ with linear probing only, while the rest of the backbone remains frozen.

The video FM $\theta_{\text{clip}}$ maps the sequence $\mathbf{x}$ to the label $y_{\text{train}}$: $\theta_{\text{clip}}: \mathbf{x}\rightarrow y$. For test sequences, the prediction is: $\hat{y}_{\text{test}} = \theta_{\text{clip}}(\mathbf{x}_{\text{test}})$.
The image-based FM $\theta_{\text{frame}}$ maps each frame $x_{i}$ to $y_{i}$: $\theta_{\text{frame}}: x_{i} \rightarrow y_{i}$. The frame fusion model $f$ combines frame-level predictions: $f: \{\theta_{\text{frame}}(x_{\text{test},i})\}_{i=1}^{t} \rightarrow y$.
The key challenge in this setting of changing camera perspectives, leads to different data distributions.

\subsection{Framework Overview}

Foundation models offer highly generalizable representations, needing minimal fine-tuning.
We use pretrained image and video foundation models as frozen backbones, trained on large datasets via self-supervised learning, acting as robust feature extractors. Linear probes fine-tune these models for fine-grained human activity recognition using the extracted features.

\begin{figure}[t]
    \centering
\includegraphics[width=1\textwidth]{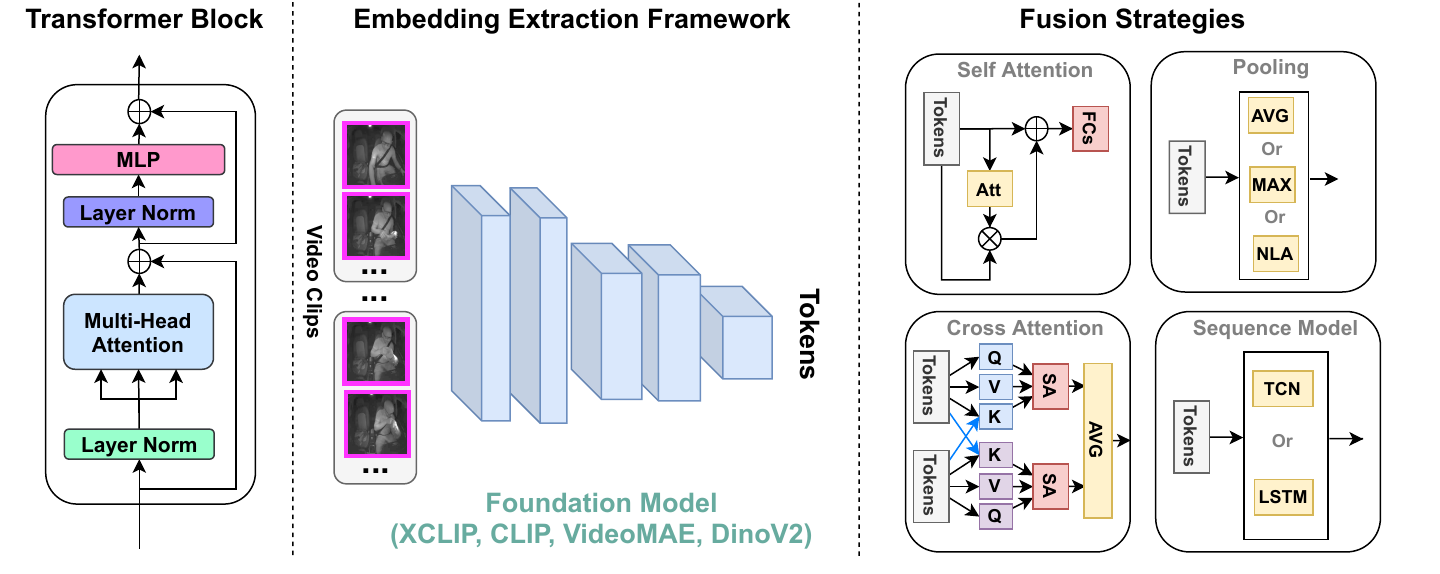}
\caption{An overview of the used framework, where the transformer block is depicted on the left hand side, the feature extraction pipeline is in the middle, and the temporal fusion strategies are illustrated on the right.}
\label{fig:main}
\vspace{-1em} 
\end{figure}

For image-based FMs $\theta_{\text{frame}}$ our framework begins with feeding input video frames through the frozen backbone to obtain frame-level representations. 
These representations are then aggregated using a temporal fusion function $f$ to form a unified video representation. For video-based FMs $\theta_{\text{clip}}$ we directly extract spatio-temporal representations of the complete input clip.

Finally, the aggregated features are passed through the linear probes, which are trained to classify the activities based on the fine-grained details captured in the video representations.
Figure \ref{fig:main} provides an overview of our framework. 

\subsection{Vision Foundation Models}

We focus vision transformers, which were proposed as general-purpose foundation models and demonstrate strong generalization performance across various tasks and domains.
Table \ref{table:FM_overview} summarizes the key details of the four foundation models examined in our study.


\subsubsection{CLIP} \cite{CLIP} uses a contrastive learning approach to align images and text in a shared embedding space. It is pretrained on a vast dataset of image-text pairs collected from the web. The model architecture consists of separate encoders for images and text, which generate embeddings for both modalities. 

\subsubsection{DinoV2} \cite{dinov2} uses self-supervised learning to train vision transformers on large-scale, diverse image datasets. The model employs a self-distillation approach without labels, where a student network learns to predict the representations generated by a teacher network. 

\begin{table}[ht]
\vspace{-1em} 
\centering
\caption{Foundation Models studied in our work.}
\label{table:FM_overview}
\begin{tabular}{cccc}
\toprule
\textbf{Foundation Model} & \textbf{Modality} & \textbf{Pretraining Data} & \textbf{Pretraining Objective} \\
\midrule
CLIP & Image + Text & WebImageText & Contrastive \\
DinoV2 & Image  & WebImage & Knowledge Distillation \\
X-CLIP & Video + Text & K400 & Multi-grained Contrastive \\
VideoMAE & Video & K400 & Masked Video Modelling \\
\bottomrule
\end{tabular}
\vspace{-1.5em} 
\end{table}

\subsubsection{X-CLIP} \cite{x-clip} is a  multi-grained contrastive model designed for video-text retrieval. It addresses a fundamental challenge in multi-modal research by incorporating cross-grained contrast, which contrasts coarse-grained and fine-grained representations. 

\subsubsection{VideoMAE} \cite{tong2022videomae} (Video Masked Autoencoders) is designed as a data-efficient learner for self-supervised video pre-training (SSVP).VideoMAE employs a customized video tube masking strategy with an extremely high masking ratio. This approach makes video reconstruction a more challenging and meaningful task, promoting the extraction of effective video representations during pre-training.

\subsection{Temporal fusion mechanism}
\label{sec:fusion}
An important area of investigation is temporal fusion in image-based foundation models used for video tasks. 
Many works follow the straight-forward approach of average pooling across frames for unified video feature representation~\cite{dinov2}. Recently, attentive probing used cross-attention on concatenated frame tokens~\cite{v-jepa}.

Inspired by these approaches, we explore various temporal aggregation methods without requiring architectural changes or Parameter Efficient Fine Tuning (PEFT). 
This section uses following notations: $T$ is the number of frames, $x_{t}$ is the feature embedding per frame and $x$ is the aggregated feature embedding.
We implement 13 different fusion mechanisms, detailed below.

\subsubsection{\underline{Pooling Methods}}
Pooling methods aggregate the features of each video frame using a combination of various pooling techniques and a non-linear function in the classification head.

\noindent \textbf{Average Pooling:} Computes the average value of features across all frames:
$
    \text{x}= \frac{1}{T} \sum_{t=1}^{T} \theta_{\text{frame}}(x_t)
$

 \noindent \textbf{Max Pooling:} Extracts the maximum value of features across each dimension of the embedding vector
$
        \text{x} = \max_{t=1}^{T}  \theta_{\text{frame}}(x_t)
$

\noindent \textbf{Pooling methods with Non-linear Activation:} To add a non-linearity to the classification head, we add a ReLU activation function in the classifier. We perform this with both Average and max pooling.

\subsubsection{\underline{Attention-Based Methods}}
In video classification tasks, attention mechanisms allow the model to focus on relevant frames or tokens within each frame. Our study explores self-attention and cross-attention, examining variations that use either all tokens or only the CLS token per frame. We introduce two notations used in this subsection:
$x_{\text{all}}$  represents all tokens per frame, while $x_{\text{cls}}$ denotes the collection consisting solely of the CLS token per frame:

\begin{equation}
\label{eq:all_tokens}
x_{\text{all}} = \left\{ x_{1}, x_{2}, \ldots, x_{T} \right\}; \quad \quad \quad x_{\text{cls}} = \left\{ x_{\text{cls},1}, x_{\text{cls},2}, \ldots, x_{\text{cls},T} \right\}
\end{equation}


\noindent\textbf{Self-Attention on all tokens w. Average Pooling} computes self-attention on  all tokens and calculates the average value of features across all frames: 
$
\text{x}= \frac{1}{T} \sum_{t=1}^{T}  \text{Self-Att} ( \theta_{\text{frame}}(x_{\text{all}}))
$

\noindent \textbf{Self-Attention on all tokens w. Max Pooling} uses self-attention on all tokens and takes the highest value across frames:
$
    \text{x}= \max_{t=1}^{T} \text{Self-Att}( \theta_{\text{frame}}(x_{\text{all}}))
$

\noindent \textbf{Self-Attention on CLS tokens w. Average Pooling} uses self-attention on the CLS tokens and averages the values across all frames:

$
    \text{x}= \frac{1}{T} \sum_{t=1}^{T} \text{Self-Att}( \theta_{\text{frame}}(x_{\text{cls}}))
$

\noindent \textbf{Self-Attention on CLS tokens w Max Pooling} uses self-attention  on CLS tokens, selecting the highest value across frames.
$
    \text{x}= \max_{t=1}^{T}  \text{Self-Att}( \theta_{\text{frame}}(x_{\text{cls}}))
$   
\noindent \textbf{Weighted Self-Attention} incorporates the attention weights from self-attention as a weighting mechanism to emphasize certain tokens more than others.

\noindent \textbf{Cross-Attention on all tokens}  is highly similar to the attentive pooling in \cite{v-jepa}. It utilizes a learnable query vector to align features from multiple frames through cross-attention \cite{crossattentionvision} on all tokens:
$    
       \text{x}= \text{Cross-Attention}( \theta_{\text{frame}}(x_{\text{all}}),q)
$

\noindent \textbf{Cross-Attention on CLS tokens} employs a learnable query vector for cross-attention \cite{crossattentionvision} specifically on CLS tokens to align features from multiple frames:
$
    \text{x}=  \text{Cross-Attention}( \theta_{\text{frame}}(x_{\text{cls}}),q)
$  

\subsubsection{\underline{Sequence Modeling Methods}} We also consider more traditional sequence modeling methods on transformer embeddings. These methods have been studied on feature maps from convolutional layers, but their effectiveness on transformer embeddings is underresearched.

\noindent \textbf{Temporal Convolution Network:} Uses dilated convolutions to capture long-range dependencies in the sequence.

\noindent \textbf{Long Short-Term Memory:} LSTMs utilize gated mechanisms to model long-term dependencies, making them adept at capturing sequential patterns over extended periods:
$  
         h_T, c_T = \text{LSTM}(x_T, h_{T-1}, c_{T-1}), 
         \quad \quad x = h_T
$

 \subsection{Testbed \& Implementation Details}



\noindent\textbf{Drive\&Act}~\cite{drive_and_act} is a dataset for fine-grained driver activity recognition in self-driving vehicles, featuring  $34$ fine-grained actions from $15$ drivers, captured using eight synchronized sensors. It includes NIR videos from six viewpoints, making it an ideal testbed for cross-view HAR. We follow the evaluation setting of~\cite{drive_and_act}, with training, validation, and test splits and no subject overlap. 
As in ~\cite{drive_and_act}, we use NIR front-top view for training and all views for evaluation.

\noindent\textbf{IKEA-ASM}~\cite{IKEA_ASM} is an extensive furniture assembly dataset with 371 unique processes involving 48 participants assembling four types of furniture in five settings. It includes 35 hours of RGB, depth, and 3D skeleton data from three perspectives, covering 33 fine-grained activities.
We follow the environment-based split~\cite{IKEA_ASM}, ensuring unique environments for training and testing. Training and validation subsets are created from three environments, while the test set includes two different environments. Our models are trained on front-view data and evaluated on side-view and top-view data to assess generalization.


\noindent\textbf{Training details:} We utilize publicly available implementations and pretrained weights for all foundational models (see Table ~\ref{table:FM_overview}). During training, we freeze the backbone of these models and train the probes using 4 Nvidia A-100 GPUs, with a batch size set to 32 over 60 epochs. We start with an initial learning rate of 1e-3 and employ the AdamW optimizer with a weight decay of 0.02 and a cosine annealing learning rate scheduler.To ensure a fair comparison with video-based foundation models, which are typically pretrained on video clips of 16 frames, we adopt 16 frames per clip as our standard during training. Spatial data augmentation is applied to generate crops sized at 224 x 224 pixels.

\noindent\textbf{Evaluation details:} We adopt standard evaluation protocols from video classification tasks \cite{vivit} by sampling three clips from each video and averaging their logits to assess performance. Following conventions in human activity recognition \cite{aganian2023object,drive_and_act,peng2022transdarc,roitberg2020cnn_spatialtemporal,coarse_net}, we use balanced accuracy (mean per class accuracy), top-1 accuracy and top-5 accuracy as our evaluation metrics. Due to the unbalanced nature of both our datasets we consider balanced accuracy to be the primary evaluation metric. In addition to testing on the same view used during training, we conduct novel view performance analysis by evaluating models on multiple unseen views. Since the number and perspectives of views vary across datasets, we compute mean novel-view balanced accuracy, top-1 accuracy, and top-5 accuracy to evaluate how effectively the models generalize to different viewpoints.
More details are provided in the supplemental material.

\section{Results and Analysis}

\subsection{Linear Probing Foundation Models for  Fine-Grained Human Activity Recognition}

First, we conduct extensive experiments using two image foundation models, CLIP \cite{CLIP} and DinoV2 \cite{dinov2}, and two video foundation models, X-CLIP \cite{x-clip} and VideoMAE \cite{tong2022videomae} with results provided in Table \ref{table:FM_eval_on Fine_grained_HAR}. 
We utilized linear probing for classification across all models and averaged the embeddings across frames specifically for the image-based models.

\begin{table}[ht]
\vspace{-1.25em} 
\centering
\caption{Performance comparison of various foundation models on fine-grained human activity recognition under trained view conditions.}
\label{table:FM_eval_on Fine_grained_HAR}
\begin{minipage}{\textwidth}
\resizebox{\textwidth}{!}{
\begin{tabular}{lc|ccc|ccc}

\hline
\toprule
\multirow{2}{*}{\textbf{Model}} & \multirow{2}{*}{\textbf{Type}} & \multicolumn{3}{c|}{\textbf{Drive\&Act}} & \multicolumn{3}{c}{\textbf{IKEA-ASM}} \\

&  & \textbf{Mean Acc} & \textbf{Top-1 Acc} & \textbf{Top-5 Acc} & \textbf{Mean Acc} & \textbf{Top-1 Acc} & \textbf{Top-5 Acc} \\

\midrule

CLIP \cite{CLIP} & Image & \cellcolor{gray!25}21.29 & 43.59 & 80.94 & \cellcolor{gray!25}6.39 & 26.47 & 71.42 \\
DinoV2 \cite{dinov2} & Image & \cellcolor{gray!25}26.99 & 50.67& 88.32 & \cellcolor{gray!25}12.27 & 22.36 & 63.27\\
X-CLIP \cite{x-clip} & Video & \cellcolor{gray!25}\textbf{40.27} & \textbf{56.86} & \textbf{90.18} & \cellcolor{gray!25}\textbf{20.87} & \textbf{37.57} & \textbf{88.12} \\
VideoMAE \cite{tong2022videomae} & Video & \cellcolor{gray!25}22.01 & 51.85 & 84.29 & \cellcolor{gray!25}5.31 & 30.97 & 63.04 \\

\bottomrule

\end{tabular}
}
\end{minipage}
\vspace{-1.25em} 
\end{table}

We first observe that image-based foundation models yield a strong performance on both datasets. DinoV2 \cite{dinov2} clearly outperforms CLIP \cite{CLIP} (7.23\% gain in Drive\&Act and 5.88\% in IKEA-ASM) in mean per-class accuracy. 
Yet, the performance on the other two metrics are not very consistent, with DinoV2 performing very well on Drive\&Act and CLIP performing well on IKEA-ASM. 
For video foundation models, X-CLIP easily outperforms VideoMAE on all metrics (mean accuracy gain of 18.26\% on Drive\&Act and 15.56\% on IKEA-ASM). 
When using simple averaging as a temporal fusion mechanism, we, therefore, cannot unanimously conclude that all video foundation models outperform image foundation models in fine-grained HAR.
While X-CLIP is the best-performing model on both datasets, VideoMAE exhibits mixed results, particularly on the IKEA-ASM dataset. We suspect that the additional modality used by X-CLIP during pretraining enhances its adaptability to fine-grained HAR.


 One observation is that the gap in performance between image and video foundation models seems to rise with the granularity of the dataset.
 For example, in Drive\&Act, the  performance gap between DinoV2 and X-CLIP is small in Top-1 Accuracy and Top-5 Accuracy, but this gap widens in the more granular IKEA-ASM dataset. The mean accuracy is always higher for X-CLIP, suggesting that image foundation models struggle to learn under-represented difficult actions compared to video models (which we will later explicitly examine in Table \ref{table:task vs fm_common_and_rare}). 


\subsection{Performance of various temporal fusion methods}
Table \ref{table:temporal_fusion_driveact} and \ref{table:temporal_fusion_ikea_asm}
 present the recognition results for various temporal fusion methods built on top of the image-based CLIP and DinoV2 as described in Section ~\ref{sec:fusion}. 
 \begin{table}[ht]
 \vspace{-1em} 
\centering
\caption{Performance comparison of various foundation models on fine-grained human activity recognition on the Drive\&Act Dataset.}
\label{table:temporal_fusion_driveact}
\begin{minipage}{\textwidth}
\resizebox{\textwidth}{!}{
\begin{tabular}{l cccc cccc}
\hline
\toprule

\multirow{1}{*}{\textbf{Temporal Fusion}} 
& \multicolumn{4}{c}{\textbf{CLIP}} & \multicolumn{4}{c}{\textbf{DinoV2}} \\

& \multicolumn{2}{c}{\cellcolor{blue!25}\textbf{Trained View}} & \multicolumn{2}{c}{\cellcolor{green!25}\textbf{Cross View}} & \multicolumn{2}{c}{\cellcolor{blue!25}\textbf{Trained View}} & \multicolumn{2}{c}{\cellcolor{green!25}\textbf{Cross View}} \\
& \textbf{Mean Acc.} & \textbf{Top-1 Acc.} & \textbf{Mean Acc.} & \textbf{Top-1 Acc.} & \textbf{Mean Acc.} & \textbf{Top-1 Acc.} & \textbf{Mean Acc.} & \textbf{Top-1 Acc.} \\

\midrule

\multicolumn{9}{l}{\textbf{Pooling Methods}} \\
Average Pooling & 21.29& 43.59& 8.58& 17.91& 26.99& 50.67& 10.58& 17.84\\
Average Pooling + ReLU & 26.82& 44.31& 10.01& 17.06& 33.28& 53.25& 10.72& 16.10\\
Max Pooling & 25.2& 49.74& 11.21& \textbf{29.62}& 31.71& 54.33& 16.06& 28.84\\
Max Pooling + ReLU & 28.93& 50.46& 13.28& 28.24& 36.96& 56.04& \textbf{17.38}& \textbf{29.65}\\
\midrule

\multicolumn{9}{l}{\textbf{Attention-Based Methods}} \\
Self-Attention -ALL +Avg. & 43.43& 61.26& \textbf{14.00}& 21.95& \textbf{56.33}& \textbf{73.5}& 15.19& 20.79\\
Self-Attention -ALL + Max. & \textbf{44.61}& \textbf{63.12}& 13.19& 22.79& 55.61& 72.01& 16.37& 20.76\\
Self-Attention -CLS +Avg. & 31.96& 52.63& 11.93& 21.49& 38.25& 59.56& 14.76& 19.41\\
Self-Attention -CLS +Max & 31.96& 53.94& 10.64& 20.48& 39.02& 59.35& 13.98& 18.58\\

Weighted Self-Attention  & 30.20& 52.27& 10.10& 19.55& 38.85& 61.21& 13.30& 19.85\\
Cross-Attention-ALL & 42.14& 59.86& 13.35& 40.12 & 54.63& 71.33& 14.88& 18.48\\
Cross-Attention-CLS & 34.12& 50.15& 11.91& 20.08& 43.06& 59.42& 14.23& 19.28\\

\midrule

\multicolumn{9}{l}{\textbf{Sequence Modeling Methods}} \\
LSTM & 31.61& 51.29& 12.34& 18.24&  41.15& 59.29& 13.88& 18.81\\
TCN & 26.56& 45.45& 7.28& 14.79& 36.01& 53.25& 9.68& 13.93\\

\bottomrule

\end{tabular}
}
\end{minipage}
\end{table}

The first interesting finding is that for simple pooling-based techniques, max pooling consistently outperforms the commonly used average pooling~\cite{dinov2} across all benchmarks and settings, with gains up to $11.71\%$ in Top-1 accuracy (CLIP in cross-view setting on Drive\&Act). Adding a ReLU activation generally improves results in most but not all cases, with the largest gain in same-view settings.
Overall, the choice of temporal fusion mechanism is critical for fine-grained HAR using foundation methods, with a performance gap of approximately $20\%$ and $10-15\%$ between average pooling and the best-performing attention-based method for Drive\&Act and IKEA-ASM (same-view setting).

Attention-based methods, especially those utilizing \textit{all tokens}, outperform pooling and sequence modeling methods for the known views. For instance, DINO v2 achieves 56.33\% mean accuracy using all tokens versus 39.02\% with the CLS token on the Drive\&Act dataset. In novel-view settings, max pooling excels, achieving 17.38\% average cross-view mean accuracy and 29.65\% top-1 accuracy for DinoV2 on Drive\&Act. Self-attention methods consistently yield the best results in fine-grained activity datasets like IKEA-ASM, with DinoV2 reaching 25.98\% average cross-view mean accuracy. Interestingly, cross-attention with all tokens (introduced as Attentive pooling in \cite{v-jepa}) is outperformed by self-attention, with a slight advantage of 2.47\% and 1.7\% in mean class accuracy for CLIP and DinoV2, respectively. Additionally, sequence modeling methods show a slight performance gain over pooling methods in the same view setting.  Overall, this outcome underscores the importance of choosing the temporal fusion mechanisms for fine-grained HAR using foundation models. A detailed discussion of dataset-specific results is provided in the supplemental material.


\begin{table}[ht]
\vspace{-1em} 
\centering
\caption{Performance comparison of various foundation models on fine-grained human activity recognition on the IKEA-ASM Dataset.}
\label{table:temporal_fusion_ikea_asm}
\begin{minipage}{\textwidth}
\resizebox{\textwidth}{!}{
\begin{tabular}{l cccc cccc}
\hline
\toprule

\multirow{1}{*}{\textbf{Temporal Fusion}} 
& \multicolumn{4}{c}{\textbf{CLIP}} & \multicolumn{4}{c}{\textbf{DinoV2}} \\

& \multicolumn{2}{c}{\cellcolor{blue!25}\textbf{Trained View}} & \multicolumn{2}{c}{\cellcolor{green!25}\textbf{Cross View}} & \multicolumn{2}{c}{\cellcolor{blue!25}\textbf{Trained View}} & \multicolumn{2}{c}{\cellcolor{green!25}\textbf{Cross View}} \\
& \textbf{Mean Acc.} & \textbf{Top-1 Acc.} & \textbf{Mean Acc.} & \textbf{Top-1 Acc.} & \textbf{Mean Acc.} & \textbf{Top-1 Acc.} & \textbf{Mean Acc.} & \textbf{Top-1 Acc.} \\

\midrule

\multicolumn{9}{l}{\textbf{Pooling Methods}} \\
Average Pooling & 6.39& 26.47& 4.46& 21.47& 12.27& 22.36& 8.55& 16.26\\
Average Pooling + ReLU & 7.38& 27.02& 4.73& 20.69& 5.08& 21.35& 3.70& 18.48\\
Max Pooling & 8.53& 32.91& 6.36& 29.46& 13.35& 37.11& 8.79& 26.4\\
Max Pooling + ReLU & 9.65& 35.48& 6.1& 28.77& 11.71& 36.33& 13.86& 32.34\\
\midrule

\multicolumn{9}{l}{\textbf{Attention-Based Methods}} \\
Self-Attention -ALL +Avg. & 20.62& 42.47& 11.86& 31.71& \textbf{25.58}& \textbf{56.52}& \textbf{21.08}& \textbf{49.22}\\
Self-Attention -ALL + Max. & \textbf{22.61}& \textbf{45.19}& \textbf{12.98}& \textbf{32.37}& 14.92& 43.4& 12.76& 39.32\\
Self-Attention -CLS +Avg. & 6.88& 28.57& 5.88& 24.49& 9.26& 32.22& 7.74& 26.59\\
Self-Attention -CLS +Max & 8.12& 28.57& 5.94& 24.52& 6.65& 26.63& 5.23& 23.76\\

Weighted Self-Attention  & 7.79& 29.27& 6.13& 25& 10.25& 33.46& 8.15& 27.65\\
Cross-Attention-ALL & 21.99& 41.77& 12.59& 21.35& 17.56& 35.56& 12.8& 28.61\\
Cross-Attention-CLS & 9.65& 30.43& 6.78& 30.05& 15.57& 33.31& 11.76& 27.21\\

\midrule

\multicolumn{9}{l}{\textbf{Sequence Modeling Methods}} \\
LSTM & 7.17& 25.23& 5.6& 21.31&  3.23& 19.41& 3.23& 19.53\\
TCN & 6.84& 27.25& 4.98& 21.35& 13.76& 35.71& 8.74& 23.06\\

\bottomrule

\end{tabular}
}
\end{minipage}
\vspace{-1em} 
\end{table}

\begin{figure}
\vspace{-1em} 
    \centering
\includegraphics[width=1\textwidth]{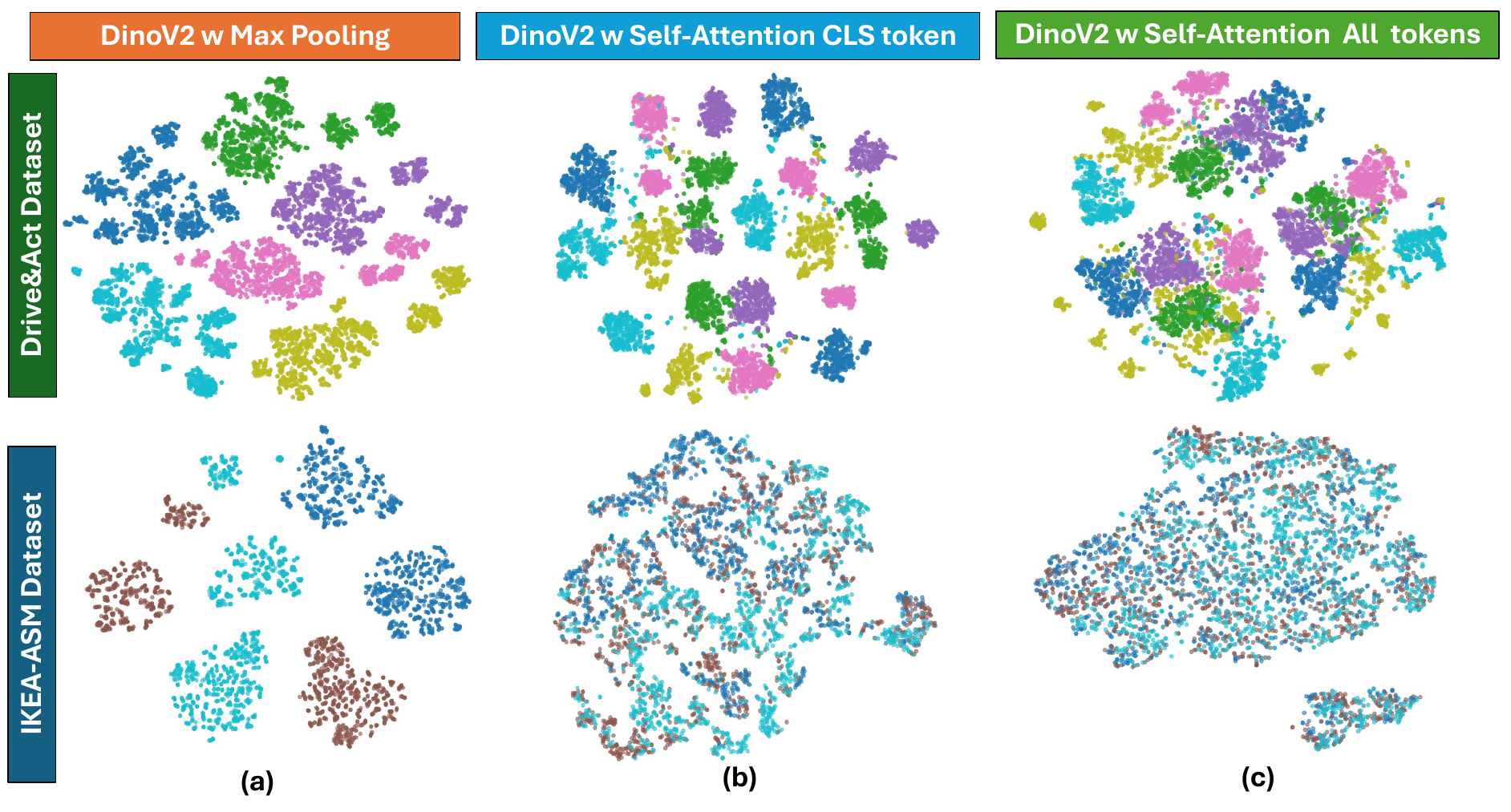}
\caption{ Visualization of embeddings on the Drive\&Act and IKEA-ASM datasets using different models. (a) DinoV2 with Max Pooling, (b) DinoV2 with Self-Attention only on CLS Token, and (c) DinoV2 with Self-Attention on All Tokens. Each color represents a different view of the dataset.}
\label{fig:t_sne_temporal}
\vspace{-1em} 
\end{figure} 

Figure \ref{fig:t_sne_temporal} also provides t-SNE visualizations of feature embeddings from DinoV2 equipped with our best temporal fusion techniques. 
Different views are clearly separated when we natively use DinoV2 (max pooling), indicating a distinct domain gap. However, when self-attention is applied, the clusters of different views come closer, suggesting that the model is learning a view-invariant representation. Ideally, models should learn a view invariant representation, where we have a cluster per class and  all views of the same class are grouped together. Our qualitative analysis demonstrates that DinoV2 with self-attention is closer to this ideal behavior, showing progress in the right direction.

\subsubsection{On video-based FMs vs. image-based FMs with temporal fusion.}
We now examine whether the temporal fusion techniques introduced in Section 5.2 enable image foundation models to surpass video models. Based on our experiments summarized in Table \ref{table:FM_eval_on_cross-viewFine_grained_HAR}, the answer is a resounding yes. In the trained-view setting, DinoV2 with self-attention on all tokens performs the best on both datasets, significantly outperforming the video-based models by 16.6\% in mean accuracy on Drive\&Act and 4.7\% on IKEA-ASM.
The novel view setting evaluation also shows that image foundation models equipped with the right fusion strategy surpass video model models, but the best-performing fusion strategy varies depending on the dataset and the backbone. There is also one exception: the video-based X-CLIP model achieves the best average cross-view mean accuracy on Drive\&Act. Yet, due to the  insignificant performance gap in this case  (17.6\% vs. 17.38\%) we view image-based FMs with temporal fusion as the more effective paradigm at the moment.

\subsection{Details on Performance in Different Novel Views}

\begin{table}[ht]
\vspace{-1em} 
\centering
\caption{Performance comparison of various foundation models on  trained view and cross view fine-grained human activity recognition on both datasets.}

\label{table:FM_eval_on_cross-viewFine_grained_HAR}
\begin{minipage}{\textwidth}
\resizebox{\textwidth}{!}{
\begin{tabular}{l cccc cccc}
\hline
\toprule

\multirow{1}{*}{\textbf{Model}} 
& \multicolumn{4}{c}{\textbf{Drive\&Act}} & \multicolumn{4}{c}{\textbf{IKEA-ASM}} \\

& \multicolumn{2}{c}{\cellcolor{blue!25}\textbf{Trained View}} & \multicolumn{2}{c}{\cellcolor{green!25}\textbf{Cross View}} & \multicolumn{2}{c}{\cellcolor{blue!25}\textbf{Trained View}} & \multicolumn{2}{c}{\cellcolor{green!25}\textbf{Cross View}} \\
& \textbf{Mean Acc.} & \textbf{Top-1 Acc.} & \textbf{Mean Acc.} & \textbf{Top-1 Acc.} & \textbf{Mean Acc.} & \textbf{Top-1 Acc.} & \textbf{Mean Acc.} & \textbf{Top-1 Acc.} \\

\midrule

\multicolumn{9}{l}{\textbf{CLIP}} \\
Average Pooling & 21.29& 43.59& 8.58& 17.91 & 6.39 & 26.47 & 4.46 & 21.47 \\
Max Pooling & 25.2& 49.74& 11.21& 29.62 & 8.53& 32.91& 6.36& 29.46 \\
Max Pooling + ReLU & 28.93& 50.46& 13.28& 28.24 & 9.65& 35.48& 6.1& 28.77 \\
Self-Attention -ALL +Avg. & 43.43& 61.26& 14.00& 21.95 & 20.62& 42.47& 11.86& 31.71 \\
Self-Attention -ALL + Max. &44.61& 63.12& 13.19& 22.79& 22.61& 45.19& 12.98& 32.37 \\
Cross-Attention -ALL  &42.14& 59.86& 13.35& 40.12&21.99& 41.77& 12.59& 21.35\\
\midrule

\multicolumn{9}{l}{\textbf{DinoV2}} \\
Average Pooling & 26.99& 50.67& 10.58& 17.84 & 12.27& 22.36& 8.55& 16.26 \\
Max Pooling & 31.71& 54.33& 16.06& 28.84 & 13.35& 37.11& 8.79& 26.4 \\
Max Pooling + ReLU & 36.96& 56.04& 17.38& \textbf{29.65} & 11.71 & 36.33 & 13.86 & 32.22 \\
Self-Attention -ALL +Avg. & \textbf{56.33}& \textbf{73.5}& 15.19& 20.79 & \textbf{25.58}& \textbf{56.52}& \textbf{21.08}& \textbf{49.22} \\
Self-Attention -ALL + Max. &  55.61& 72.01& 16.37& 20.76 & 14.92& 43.4& 12.76& 39.32 \\
Cross-Attention -ALL  &54.63& 71.33& 14.88& 18.48& 17.56& 35.56& 12.8& 28.61 \\
\midrule

\multicolumn{9}{l}{\textbf{Video Foundation Models}} \\
X-CLIP & 40.27 & 56.86 & \textbf{17.60} & 25.67 &  20.87 & 37.57 & 11.23 & 29.19 \\
VideoMAE & 22.01 & 51.85 & 11.70 & 25.23 & 5.31 & 30.97 & 3.95 & 23.33 \\
\midrule
\multicolumn{9}{l}{\textbf{Random Baseline}} \\
Random & 2.94 & 2.94 & 2.94 & 2.94 & 3.03 & 3.03 & 3.03 & 3.03 \\

\bottomrule

\end{tabular}
}
\end{minipage}
\end{table}
\begin{figure}[ht]
\vspace{-1.5em} 
    \centering
\includegraphics[width=0.8\textwidth]{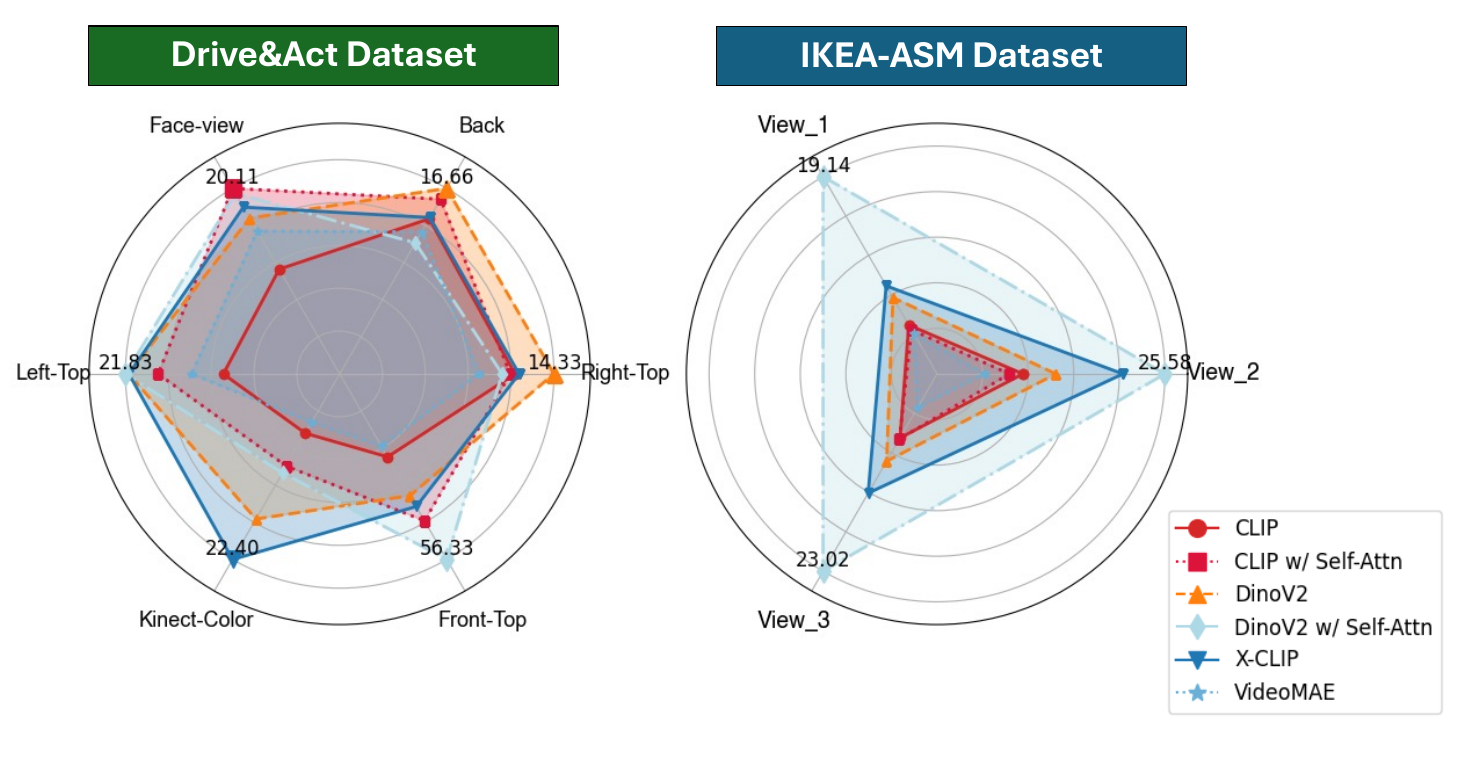}
\caption{Performance on the individual views. The plots show the mean class accuracy of various foundation models in both the trained and novel view settings}
\label{fig:cross_view_metrics}
\vspace{-1em} 
\end{figure}

\begin{figure}[ht]
\vspace{-0.5em} 
    \centering
\includegraphics[width=1\textwidth]{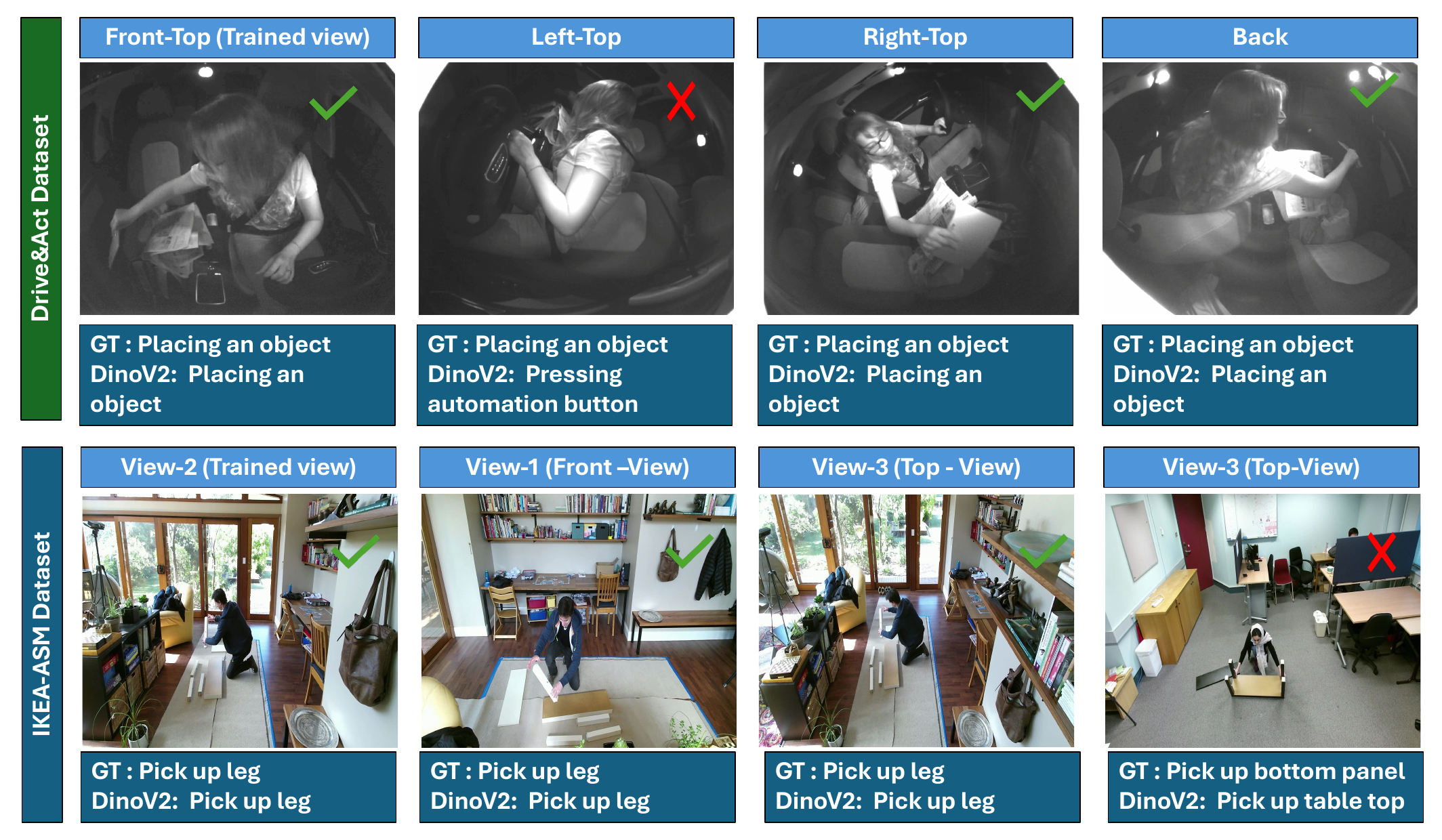} 
\caption{Qualitative results of DinoV2 with Self-attention fusion on different views as well as the Ground Truth (GT) activity. Green check marks and red crosses indicate correct and incorrect predictions, respectively. 
}
\label{fig:model_outputs}
\vspace{-2em} 
\end{figure} 


In Figure \ref{fig:cross_view_metrics} we provide recognition results for the individual novel views. 
In the Drive\&Act benchmark, some models perform well in specific views, while IKEA-ASM shows consistent performance with DinoV2 and Self Attention excelling in all views. Figure \ref{fig:model_outputs} illustrates our model outputs across different views of the same clip. It is clear that models struggle with occluded objects (e.g., the left-top view in the top row) or similar objects (e.g., bottom panel and table top in the bottom row). 

\subsection{Task-Specific vs. Foundation Models}
Finally, we analyze how foundation models compare against task-specific models like I3D \cite{qiu2017learning} and Video Swin \cite{liu2021video_swin}. 
Following the settings of ~\cite{drive_and_act,roitberg2020cnn_spatialtemporal}, I3D weights are initialised with the Kinetics~\cite{carreira2017quo} dataset, while Video Swin is pretrained on Kinetics 400~\cite{kay2017kinetics400}.
Note, that the task-specific models are \textit{fully fine-tuned}, while we only use linear probing on the last layer for foundation models. 

Table \ref{table:task vs fm in drive_and act} compares the performance of these models against our best image and video foundation models.

\begin{table}[b]
\vspace{-1.25em} 
\centering
\caption{Task-specific vs. foundation models across various views of  Drive\&Act.}
\label{table:task vs fm in drive_and act}
\begin{minipage}{\textwidth}
\resizebox{\textwidth}{!}{
\centering
\begin{tabular}{lcccccccccccc}
\toprule
\textbf{Model} & \multicolumn{2}{c}{\textbf{Trained-view}} & \multicolumn{2}{c}{\textbf{Right-top}} & \multicolumn{2}{c}{\textbf{Back}} & \multicolumn{2}{c}{\textbf{Face-view}} & \multicolumn{2}{c}{\textbf{Left-Top}} & \multicolumn{2}{c}{\textbf{Kinect-color}} \\
 & {\cellcolor{orange!25}\textbf{Acc.}} & {\cellcolor{teal!25}\textbf{Mean  Acc.}} & {\cellcolor{orange!25}\textbf{Acc.}} & {\cellcolor{teal!25}\textbf{Mean  Acc.}} & {\cellcolor{orange!25}\textbf{Acc.}} & {\cellcolor{teal!25}\textbf{Mean  Acc.}} & {\cellcolor{orange!25}\textbf{Acc.}} & {\cellcolor{teal!25}\textbf{Mean  Acc.}} & {\cellcolor{orange!25}\textbf{Acc.}}& {\cellcolor{teal!25}\textbf{Mean  Acc.}} & {\cellcolor{orange!25}\textbf{Acc.}} & {\cellcolor{teal!25}\textbf{Mean  Acc.}} \\
\midrule
I3D  & 63.64 & - & 4.51 & - & 6.96 & - & 7.39 & - & 9.03 & - & 5.41 & - \\
Video Swin & 72.88 & \textbf{58.98} & 26.81 & \textbf{19.77} & \textbf{26.7} & \text{19.61} & 25.52 & \textbf{19.92} & 31.04 & \textbf{27.29} & 23.61 & 10.33 \\
DinoV2 w Max Pooling & 56.04 & 36.96 & \textbf{30.94} & 14.33 & 22.57 & 16.66 & \textbf{28.41} & 16.89 & \textbf{32.23} & 21.51 & \textbf{34.09} & 17.52 \\
DinoV2 w Self Attention & \textbf{73.5} & 56.63 & 16.74 & 10.87 & 16.73 & 11.77 & 25.41 & 19.55& 25.21 & 21.83 & 19.88 & 11.95 \\
X-CLIP & 56.86 & 40.27 & 24.01 & 12 & 21.95 & 14.04 & 26.86 & 18.08 & 27.53 & 21.48 & 27.99 & \textbf{22.4} \\
\bottomrule
\end{tabular}
}
\end{minipage}
\vspace{-1.25em} 
\end{table}

There are several insights: DinoV2 with max pooling outperforms task-specific models in novel views for Top-1 Accuracy, while DinoV2 with self-attention excels in trained views in the Drive\&Act dataset. However, Video Swin has the highest mean accuracy in both trained and novel views. This discrepancy likely arises because foundation models only fine-tune probes, struggling with underrepresented classes.

For a deeper analysis in Table \ref{table:task vs fm_common_and_rare}, we follow the evaluation protocol of~\cite{roitberg2020cnn_spatialtemporal}, sorting the action classes by frequency and splitting them into common (first half) and rare (second half).
\begin{table}
\vspace{-1.25em}
\centering
 
\caption{Task-specific models vs. foundation models evaluated separately for common and rare activity classes of the Drive\&Act dataset.}
\label{table:task vs fm_common_and_rare}
\begin{minipage}{\textwidth}
\resizebox{\textwidth}{!}{
\centering
\begin{tabular}{lcccccccc}
\toprule
\textbf{Model} & \multicolumn{4}{c}{{\cellcolor{blue!25}\textbf{Trained View}}} & \multicolumn{4}{c}{{\cellcolor{green!25}\textbf{Cross View}}} \\
 & \multicolumn{2}{c}{\textbf{Common Classes}} & \multicolumn{2}{c}{\textbf{Rare Classes}} & \multicolumn{2}{c}{\textbf{Common Classes}} & \multicolumn{2}{c}{\textbf{Rare Classes}} \\
 & {\cellcolor{orange!25}\textbf{Acc.}} & {\cellcolor{teal!25}\textbf{Mean Acc.}} & {\cellcolor{orange!25}\textbf{Acc.}} & {\cellcolor{teal!25}\textbf{Mean Acc.}} & {\cellcolor{orange!25}\textbf{Acc.}} & {\cellcolor{teal!25}\textbf{Mean Acc.}} & {\cellcolor{orange!25}\textbf{Acc.}} & {\cellcolor{teal!25}\textbf{Mean Acc.}} \\
\midrule
Video Swin & 75.85 & 66.48 & \textbf{56.54} & \textbf{51.11} & 29.35 & \textbf{24.53} & 13.12 & \textbf{14.45} \\
DinoV2 w Max Pooling & 59.48 & 36.55 & 27.56 & 26.86 & \textbf{31.89} & 18.87 & 13.00 & 13.25 \\
DinoV2 w Self Attention & \textbf{77.83} & \textbf{67.91} & 50.96 & 44.75 & 22.47 & 18.02 & 12.11 & 12.36 \\
X-CLIP & 61.6 & 52.32& 32.15 & 28.22 & 28.05 & 21.99 & \textbf{13.28} & 13.20 \\
\bottomrule
\end{tabular}
}
\end{minipage}
\vspace{-1.25em} 
\end{table}
In the trained view, DinoV2 with self-attention surpasses Video Swin by 1.98\% in Top-1 Accuracy and 1.43\% in mean accuracy for common classes. However, Video Swin significantly outperforms DinoV2 in rare classes by 5.58\% in Top-1 Accuracy and 6.36\% in mean accuracy which provides evidence to our earlier statement. Interestingly, in the cross-view setting, Video Swin maintains better mean accuracy for both common and rare classes. We suspect this is because we consider average metrics across views, and the performance of models fluctuates across views, obscuring a clear picture. This explains why the performance in rare classes is very close for all these models.

\section{Conclusion}
We presented an exploratory study on fine-grained human activity understanding and cross-view generalization of pre-trained and minimally fine-tuned foundation models. 
We examined different variants of four pre-trained foundation models, including two video-based and two image-based architectures, along with 13 temporal fusion mechanisms. Our findings indicate that DinoV2 is particularly effective in this setting, and the temporal fusion strategy significantly impacts the outcome. Notably, max pooling outperformed the conventionally used frame averaging, while the best results are mostly achieved with self-attention with all tokens.
Our experiments provide encouraging evidence that modern foundation models could benefit from more advanced temporal modeling methods.


%
%
%
%
\bibliographystyle{splncs04}
\bibliography{main}

\end{document}